\documentclass[sigconf]{acmart}
\usepackage[plain]{fancyref}

\usepackage{amsmath}
\usepackage{graphicx,psfrag,epsf}
\usepackage{enumerate}
\usepackage{url} 
\usepackage{helvet}
\usepackage{amssymb}
\usepackage{amsthm}
\usepackage{lineno}
\usepackage{url}
\usepackage{bm}
\usepackage{algorithm, algorithmic}
\usepackage{float}
\usepackage[toc,page]{appendix}
\usepackage{setspace}
\usepackage{multirow} 
\usepackage{bbm}
\usepackage{enumitem} 
\usepackage{indentfirst}
\usepackage{comment}
\usepackage{xcolor}
\usepackage{breakurl}

\newcommand{\algnameRM}{\textsc{RM}}
\newcommand{\algnamePDD}{\textsc{PDD}}

\newcommand{\algnameNIQCD}{\textsc{NIQCD}}

\newcommand{\Tra}{^{\sf T}} 

\newcommand{\V}[1]{{\bm{\mathbf{\MakeLowercase{#1}}}}} 
\newcommand{\M}[1]{{\bm{\mathbf{\MakeUppercase{#1}}}}} 

\AtBeginDocument{%
  \providecommand\BibTeX{{%
    \normalfont B\kern-0.5em{\scshape i\kern-0.25em b}\kern-0.8em\TeX}}}

\copyrightyear{2020} 
\acmYear{2020} 
\setcopyright{acmlicensed}
\acmConference[KDD '20]{Proceedings of the 26th ACM SIGKDD Conference on Knowledge Discovery and Data Mining}{August 23--27, 2020}{Virtual Event, CA, USA}
\acmBooktitle{Proceedings of the 26th ACM SIGKDD Conference on Knowledge Discovery and Data Mining (KDD '20), August 23--27, 2020, Virtual Event, CA, USA}
\acmPrice{15.00}
\acmDOI{10.1145/3394486.3403240}
\acmISBN{978-1-4503-7998-4/20/08}

\begin{document}
\fancyhead{}

\title{A Non-Iterative Quantile Change Detection Method in Mixture Model with Heavy-Tailed Components}

\author{Yuantong Li}
\email{li3551@purdue.edu}
\affiliation{%
  \institution{Department of Statistics, Purdue University}
}

\author{Qi Ma}
\email{qma4@ncsu.edu}
\affiliation{%
  \institution{Department of Statistics, North Carolina State University}
}

\author{Sujit K. Ghosh}
\email{sujit.ghosh@ncsu.edu}
\affiliation{%
  \institution{Department of Statistics, North Carolina State University}
}

\begin{abstract}
 Estimating parameters of mixture model has wide applications ranging from classification problems to estimating of complex distributions. Most of the current literature on estimating the parameters of the mixture densities are based on iterative Expectation Maximization (EM) type algorithms which require the use of either taking expectations over the latent label variables or generating samples from the conditional distribution of such latent labels using the Bayes rule. Moreover, when the number of components is unknown, the problem becomes computationally more demanding due to well-known label switching issues \cite{richardson1997bayesian}. In this paper, we propose a robust and quick approach based on change-point methods to determine the number of mixture components that works for almost any location-scale families even when the components are heavy tailed (e.g., Cauchy). We present several numerical illustrations by comparing our method with some of popular methods available in the literature using simulated data and real case studies. The proposed method is shown be as much as 500 times faster than some of the competing methods and  are also shown to be more accurate in estimating the mixture distributions by goodness-of-fit tests.
\end{abstract}



\maketitle

\section{Introduction}

\label{sec:intro}
Determining the number of components in a finite mixture model is crucial in many application areas such as financial data \cite{hu2006dependence,wong2009student, si2013exploiting}, bio-medical studies \cite{humphrey2002constrained,yin2008medical} and low-frequency accident occurrence prediction \cite{wang2011predicting,park2014finite}. Existing literature have witnessed numerous computational methods, and in particular Markov Chain Monte Carlo methods \cite{geweke2007interpretation,carlin1995bayesian,wang2016hybrid} and EM algorithms \cite{mccallum1999multi, muthen1999finite, ma2017feature} have been used with a lot of success. However, either these methods are computationally demanding and/or these methods are developed under the assumption of data being generated from mixtures of densities from the exponential family, in part because the family of exponential distribution has a sufficient statistic of constant dimension (i.e., the dimension of the sufficient statistic remains fixed for any sample size) and so the updates of the data augmentation type algorithm involve their smaller dimensional sufficient statistics \cite{fearnhead2005direct,fraser1963sufficiency,neal2015exact}. One the other hand, many heavy-tailed distributions (e.g., Cauchy, t etc.) do not have such nice properties and Gibbs or EM updates become much more involved and computationally demanding \cite{geman1984stochastic}. However, we cannot always assume data being generated by a mixture of distributions with exponential tails which necessarily makes the mixture to be exponentially tailed. For example, financial data (e.g., log-returns of stock prices) often exhibit higher peaks and heavy tails \cite{grossman1980determinants} and it would be challenging for traditional MCMC or EM algorithms to detect the number of components if we fit a mixture model to such data.

The above-mentioned methods, including the EM algorithm \cite{dempster1977maximum}, determine the number of components in the mixture model by using non-nested statistical test and its variants. For example, \cite{zhang2004determine} provided an extended KS test based on the EM procedure to determine the number of components in parallel. \cite{benaglia2011bandwidth} provided a new selecting bandwidth method in kernel density estimation used by the nonparametric EM (npEM) algorithm of \cite{benaglia2009like}. Besides, \cite{roeder1994graphical} used the diagnostic plot to estimate the number of components under the assumption that components are normally distributed with a common variance and there are only a few number of components. In recent years, \cite{rousseau2011asymptotic} (RM) have demonstrated the asymptotic behavior of the number of components in overfitted model which tends  to zero-out the extra components, and this stability can be realized by imposing restrictions on the prior. \cite{nasserinejad2017comparison} illustrated the success of the RM method by applying it to hemoglobin values of blood donors. 

More commonly, information criteria like AIC \cite{akaike1987factor}, BIC \cite{burnham2004multimodel} and DIC \cite{ward2008review} have also been used for model selection for mixture models despite the fact that mixture models lead to so-called singular models that violates the standard regularity conditions required for the consistency of the criteria. \cite{drton2017bayesian} proposed a modified singular Bayesian information criterion (sBIC) for models in which the Fisher information matrices may not be strictly positive definite (leading to singular model). However, the sBIC is difficult to construct especially when the components of the mixture models are assumed to arise from non-exponential families.

In this paper, we developed an easily implementable data-driven method that we call the Non-Iterative Quantile Change Detection (NIQCD) by using change-point detection methods applied to the quantiles of the component distributions of the mixture model. We adapt the RM \cite{rousseau2011asymptotic} framework by starting with a relatively larger number of components (relative to the sample size) and empirically estimate the location and scale parameters of the components using quantiles conditioned on the subset of data assumed to be obtained from the corresponding components. The probability (mixing) weights assigned to each component are then estimated by constrained least squares method making the method computationally much faster. We use the change-point method \cite{killick2012optimal} to reduce the redundancy of the location parameter and shrink it to a stable state. To summarize the three main highlights of the data-driven NIQCD are: (1) the quantile based estimation can be applied to any location-scale family (including heavy-tailed distributions); (2) the constrained least square method for estimating the weights makes the algorithm non-iterative; and finally (3) the change-point based method reduces the dimensionality of the mixture model to an optimal number. 

We illustrate the performance of two methods as compared to NIQCD for estimating the number of components for a mixture model with Cauchy components. It is also to be noticed that compared to distributions whose tails decay exponentially, Cauchy distribution is much heavy-tailed, which makes the detection of hidden components overlap much harder (see Section 2 for further details on this aspect). In our simulation studies we have not used certain Bayesian techniques, such as Reversible-jump Markov chain Monte Carlo \cite{richardson1997bayesian} which we found computationally much slower compared to other methods that we have chosen to compare with the proposed NIQCD in terms of computational time and estimation accuracy. In particular we consider two Bayesian model selection methods in our experiment: RM \cite{rousseau2011asymptotic}, posterior deviance distribution (PDD) \cite{aitkin2015new}, and our proposed baseline model iterative QCD (IQCD).

The rest of this paper is organized as follows. In Section 2, we provide the  general definition of mixture model. In Section 3, we give two definitions of measuring the overlap between multiple distributions. In Section 4, we describe two algorithms IQCD and NIQCD. In section 5, related literature is summarized. In Section 6, we present the simulation design and results for comparing R\&M, PDD, IQCD, and NIQCD. In Section 7, we applied NIQCD to stock S\&P 500 index return data. Finally, we discuss potential extensions of NIQCD.

\section{Objectives and Notations}
A $m$-component mixture of location-scale families has the following density:
\begin{equation}
f_m(x; \bm{\theta}) = \sum_{k=1}^m\lambda_k g\left({\frac{x-\mu_k}{\sigma_k}}\right){\frac{1}{\sigma_k}},
\label{eq:mixmodel}
\end{equation}
where $g(\cdot)$ is a known unimodal density from location-scale families, $\bm{\mu}=(\mu_1,\ldots,\mu_m)\in\mathbb{R}^m, \bm{\sigma}=(\sigma_1,\ldots,\sigma_m)\in (0, \infty)^m$, $\bm{\lambda}=(\lambda_1,\ldots,\lambda_m) \in {\mathcal{S}}_m$, ${\mathcal{S}}_m=\{(\lambda_1,\ldots,\lambda_m): \lambda_k \geq 0, \sum_{k=1}^{m} \lambda_k=1\}$ denotes the $m$-dimensional simplex space, and $\bm{\theta} = (\bm{\mu}, \bm{\sigma}, \bm{\lambda})$.

Given a sample of observations $X_i\stackrel{iid}{\sim} f_{m}(\cdot)$ for $i=1,\ldots,n$, the empirical cumulative distribution function (ecdf) is defined as $F_n(y) = \frac{1}{n}\sum_{i = 1}^{n} I(x_{i} \leq y)$. our goal is to estimate the true number of mixture components $m$. As a secondary interest, we also want to estimate $\bm{\theta} = (\bm{\mu}, \bm{\sigma}, \bm{\lambda})$ corresponding to the estimated value of $m$.

\section{Measure of overlapping and dispersion across components}
In this section, we develop formal concepts of the degree of the overlapping and the amount of dispersion across different components. We use these concepts to build test cases for our simulation studies in Section 6.

When determining the number of components in a mixture model, the amount of difficulty often depends on the degree of overlapping (DOL) and between component dispersion (BCD). If the components have non-overlapping supports and large dispersion, it is generally easy to detect the number of components. However, if the components have heavy degree of overlapping supports and small dispersion, the detection of the number of components becomes harder (at least visually). When focusing on unimodal location-scale families, such DOL and BCD are manifested by location parameters. For example, if $\mu_k=\mu$ for all $k$, the mixture density is unimodal and hence visually it is often infeasible to identify the mixture components and hence makes the estimation of $m$ more difficult \cite{nowakowska2014tractable}.

Below we present two criteria for a general class of mixture models which is not limited to location-scale components.


\subsection{Weighted Degree of Overlapping}
Consider the case of two components ($m=2$) in a mixture model, that is, $f(x)=\lambda_1g_1(x)+\lambda_2g_2(x)$, where $\lambda_1+\lambda_2=1$, and $g_1(x), g_2(x)$ are arbitrary densities. A measure of overlapping between $g_1$ and $g_2$ can be defined as:
\begin{equation}
\text{DOL} = \int \min(g_1(x),g_2(x)) dx=1- \frac{1}{2} \int |g_1(x)-g_2(x)| dx,
\end{equation}
where $|g_1(x)-g_2(x)|=g_1(x)+g_2(x)-2\min(g_1(x),g_2(x))$. 

Clearly, $0\leq\int \min(g_1(y),g_2(y)) dy\leq 1$ and the boundary values $0$ and $1$ are achieved when $g_1$ and $g_2$ have disjoint supports (i.e., no overlapping) and $g_1=g_2$ (i.e., complete overlapping), respectively. However, such widely used measure of overlapping doesn't take into account the weights assigned to each component and its generalization to $m(>2)$ can be difficult.

With above limitations, we propose weighted degree of overlapping (wDOL) for a mixture model as

\begin{equation}
\text{wDOL} = \frac{\int \underset{1\leq k\leq m}{\min}\; \lambda_kg_k(x)dx}{\underset{1\leq k\leq m}{\min}\; \lambda_k},
\end{equation}
where $g_k(x)$ is the $k$th component and $\lambda_k$ is the corresponding weight for $k=1,2,...,m$.

Without loss of generality,  we assume $\lambda_1\geq \ldots \geq \lambda_m > 0$. Since $0\leq\min_k\{\lambda_k g_k(x)\}\leq \lambda_kg_k(x)$ for arbitrary $x$ and each $k$, we have

\begin{equation}
0 \leq \text{wDOL}\leq \frac{\lambda_m \int g_m(x)dx}{\lambda_m}=1,
\end{equation}
where the lower boundary is achieved when $\min_k\{\lambda_k g_k(x)\}=0$ for almost all $x$, i.e., when the intersection of the supports of $\{g_k\}_{k=1}^{m}$ is a null set, and the upper boundary is achieved when all components are all identical.

The well-defined wDOL also possesses the desired property that it is invariant under a class of transformation:

\begin{equation}
\frac{\int \underset{1\leq k\leq m}{\min}\; \lambda_kg_k(x)dx}{\underset{1\leq k\leq m}{\min}\; \lambda_k} = \frac{\int \underset{1\leq k\leq m}{\min}\; \lambda_kg_k(T(x))|dT(x)|}{\underset{1\leq k\leq m}{\min}\; \lambda_k},
\end{equation}
where $T(\cdot)$ is a strictly monotone differentiable function.

\subsection{Robust Between Component Dispersion}

The mixture model can also be expressed in a hierarchical form. That is, $X \sim \sum_{k=1}^m\lambda_kg_k(x)$ is equivalent to $X|Z=k\sim g_k(x)$, where $P[Z=k]=\lambda_k, k=1,2,\ldots, m$. Thus, the between component dispersion (BCD) can be defined as
\begin{equation}
\text{BCD} = \frac{\text{Var}(\mathbb{E}(X|Z))}{\text{Var}(X)},
\end{equation}
where $\mathbb{E}(X^2|Z=k)<\infty, \forall k$.

Since $\text{Var}(X)=\mathbb{E}(\text{Var}(X|Z)) + \text{Var}(\mathbb{E}(X|Z))$, BCD $\in[0,1]$. The lower boundary is achieved when components' location parameters are identical, whereas the upper boundary is achieved when there is almost no dispersion among components. However, such measure of between component dispersion doesn't extend to the situation where the components are heavy-tailed densities (e.g., Cauchy distribution that doesn't have the second moment).

Therefore, we propose a robust version of the between component dispersion (rBCD) using more robust measures of variability:

\begin{equation}
\text{rBCD} = \frac{\mathbb{E}|\text{Med}(X|Z) - \text{Med}(X)|}{\mathbb{E}|X-\text{Med}(X)|}. 
\end{equation}

Note that the median of $X$ denoted by $\text{Med}(X)$ is always well defined and minimizes $g(\mu)=\mathbb{E}[|X-\mu|-|X|]$. Thus, we have $\mathbb{E}[|X-\text{Med}(X)|-|X|]\leq \mathbb{E}[|X-\text{Med}(X|Z)|-|X|]+\mathbb{E}|\text{Med}(X|Z)-\text{Med}(X)|$ because $|x-\mu|-|x|\leq |\mu|$. It follows that rBCD $\approx 0$ when the dispersion between components is large, whereas rBCD $\approx 1$ when the dispersion between components is small.

\section{Estimation using Quantile Change Detection (QCD) Method}


In this section, we propose two methods to determine the number of components and estimate the parameters in the mixture model whose components are from the exponential family. The \textit{Iterative Quantile Change detection} method (IQCD) is based on an iterative procedure to update the weight parameter and estimate the number of true latent classes by excluding tiny weight components. The \textit{Non-Iterative Quantile Change Detection} method (NIQCD) determines the number of components by solving the change-point problem and estimates the parameters by using empirical cumulative distribution function (eCDF) to build up the linear equations. The coordinate descent method can also be integrated into NIQCD to refine the estimates of the location, scale, and weight parameters but with additional computational time.


For identifiability of the components, we further assume that $\mu_{1} \leq \mu_{2} \leq \ldots \leq \mu_{m}$.

\subsection{Iterative QCD (IQCD)}

The idea of IQCD is to select relatively large weights and use the cumulative sum criteria (\textit{cusum}) \cite{page1954continuous} to select the corresponding components followed by updating the parameters. By iterating this procedure, we could finally converge to a stable weight distribution and determine the number of components. The detailed steps are described as follows.

\begin{enumerate}[leftmargin=1.5em]
\item \textbf{Initialize $m$ and $\bm{\mu}$}. 
Rough estimates of $m$ and $\bm{\mu}$ can be obtained by setting $\hat{m}^{(\text{init})}=[\sqrt{n}]$ and
\begin{equation}
\hat{\mu}_k^{(\text{init})} = x_{([\frac{nk}{\hat{m}^{(\text{init})}+1}])}, k=1,2,\ldots,\hat{m}^{(\text{init})}
\end{equation}
where $[\cdot]$ takes the integer value, and $\{x_{(i)}\}_{i=1}^n$ are the ordered statistics of the samples.

Then the `change point' \cite{ebrahimi2001bayesian,lee2010change, killick2012optimal} method is used to identify the significant difference of the sudden change  between those $\hat{\mu}_k^{(\text{init})}$, which is implemented by minimizing a cost function to detect change points. Then we obtain the estimated number of component $\hat{m}^{(0)}$ usually less than $\hat{m}^{(\text{init)}}$, and re-estimate $\hat{\mu}_k^{(0)}$ based on $\hat{m}^{(0)}$, where $k=1,\ldots,\hat{m}^{(0)}$.

\item \textbf{Initialize $\bm{\sigma}$}.
Notice that for the standard Cauchy, the CDF is $G(z) = 1/2 + (1/\pi) \tan^{-1}(z)$, and $G(1) = 3/4$. For the $k$-th component, we want $(\mu_{k} + \mu_{k+1})/2$ as the third quartile in the standard Cauchy distribution. So we get $G((\frac{\mu_{k} + \mu_{k+1}}{2} -\mu_{k})/\sigma_{k}) = G((\mu_{k+1} - \mu_{k})/2\sigma_{k}) = 3/4$, leading to $(\mu_{k+1}-\mu_{k})/2\sigma_{k} = 1$. Therefore, the scale parameters $\sigma_{k}, k = 1, \ldots, \hat{m}$, can be estimated by quantiles of general Cauchy distribution
\begin{equation}
    \begin{aligned}
        \hat{\sigma}_{k}^{(0)} = \frac{1}{2\tau} &
        \left[(\frac{k+1}{\hat{m}^{(0)}+2}) \text{th} \enskip \text{quantile of data}\right] \\
        & - \frac{1}{2\tau} \left[ (\frac{k}{\hat{m}^{(0)}+2}) \text{th} \enskip \text{quantile of data}\right].
    \end{aligned}
\end{equation}
So $\hat{\sigma}_{k}^{(0)} = [x_{([\frac{n(k+1)}{\hat{m}^{(0)}+2}])} - x_{([\frac{nk}{\hat{m}^{(0)}+2}])}]/(2\tau)$ and $\tau \geq 1$ which represents the hyperparameter and can be adjusted by the needs for more general conditions.\footnote{Default $\tau = 1$ if Cauchy distribution.}

\item \textbf{Initialize} $\bm{\lambda}$.
For the $k$-th component, we simply use the difference of eCDF between two adjacent components, $k-1$ and $k+1$, to represent the weight for the $k$-th component,
\begin{equation}
    \begin{aligned}
        \hat{\lambda}_{k}^{(0)} = F_n(\frac{\hat{\mu}_{(k+1)}^{(0)}+\hat{\mu}_{(k)}^{(0)}}{2})
        - F_n(\frac{\hat{\mu}_{(k)}^{(0)}+\hat{\mu}_{(k-1)}^{(0)}}{2}), 
    \end{aligned}
\end{equation}
where $k = 2,\ldots, \hat{m}^{(0)}-1$, $\hat{\lambda}_{1}^{(0)} = F_n((\hat{\mu}_{(2)}^{(0)}+\hat{\mu}_{(1)}^{(0)})/2) - F_{n}(x_{(1)})$ corresponds to the first component, and $\hat{\lambda}_{\hat{m}^{(0)}}^{(0)} =  F_{n}(x_{(n)}) - F_n((\hat{\mu}_{(\hat{m}^{(0)})}^{(0)}+\hat{\mu}_{(\hat{m}^{(0)}-1)}^{(0)})/2)$ is for the last component.

\item \textbf{Initialize latent variables}. 
For each data point $x_{i}$, it has a corresponding probability of belonging to the $k$th component $p_{i,k} = \mathbb{P}[Z_{i} = k]$, which can be initialized using
\begin{equation}
\begin{aligned}
\mathbb{P}[Z_{i}^{(0)} = k] = \hat{p}_{i,k}^{(0)} = \frac{\hat{\lambda}_{k}^{(0)}\frac{1}{\sigma_{k}^{(0)}} g(\frac{x_i - \hat{\mu}_{k}^{(0)}}{\sigma_{k}^{(0)}})} {\sum_{k = 1}^{\hat{m}^{(0)}} \hat{\lambda}_{k}^{(0)}\frac{1}{\sigma_{k}^{(0)}} g(\frac{x_i - \hat{\mu}_{k}^{(0)}}{\sigma_{k}^{(0)}})}.
\end{aligned}
\end{equation}
It's easy to show that $\sum_{k = 1}^{\hat{m}^{(0)}}p_{i,k} = 1, \forall i$.

\item \textbf{Update parameters iteratively}. 
In the $l$-th iteration, we update the parameters as follows:
\begin{equation}
\begin{aligned}
\hat{\mu}_{k}^{(l)} &= \text{Med}\{x_i:  Z_{i}^{l-1} = k\}, \\
\hat{\sigma}_{k}^{(l)} &= \text{IQR}\{x_i: Z_{i}^{l-1} = k\}, \\
\hat{\lambda}_{k}^{(l)} &= \frac{1}{n} \sum_{i = 1}^{n} \hat{p}_{i,k}^{(l-1)},
\end{aligned}
\end{equation}
\begin{equation}
    \begin{aligned}
        \mathbb{P}[Z_{i}^{(l)} = k] &= \hat{p}_{i,k}^{(l)} =  \frac{\hat{\lambda}_{k}^{(l)}\frac{1}{\sigma_{k}^{(l)}} g(\frac{x_i - \hat{\mu}_{k}^{(l)}}{\sigma_{k}^{(l)}})} 
{\sum_{k = 1}^{\hat{m}^{(0)}} \hat{\lambda}_{k}^{(l)}\frac{1}{\sigma_{k}^{(l)}} g(\frac{x_i - \hat{\mu}_{k}^{(l)}}{\sigma_{k}^{(l)}})},
    \end{aligned}
\end{equation}
where \textit{Med} is the median and \textit{IQR} is the interquartile range.

Based on cusum criteria\footnote{Cusum criteria: It uses the cumulative sum of estimated weights. One could obtain a distribution of $\hat{m}^{(l)}$ over iteration if the accumulated sum of ordered weights is larger than a certain threshold e.g. $1-\epsilon$. In other words, the (assumed) true number of non-empty classes ($m$) of iteration $l$ could be computed in each MCMC iteration as:
$\hat{m}^{(l)} = \min k; \; \text{s.t.} \sum_{j=1}^{k} \hat{\lambda}_{(j)}^{(l)} \geq 1- \epsilon$, where $\hat{\lambda}_{(j)}^{(l)}$ is the iteration $l$'s $(j)$th large weight and $\hat{\lambda}_{(1)}^{(l)} \geq \hat{\lambda}_{(2)}^{(l)}, \ldots, \geq \hat{\lambda}_{(\hat{m}^{(l-1)})}^{(l)}$. We usually can set $\epsilon = 0.01, 0.05$ or $0.1$. }, the estimated component $\hat{m}^{(l)}$ for at iteration $l$ is
\begin{equation}
\hat{m}^{(l)} =  \min k; \; \text{s.t.} \sum_{j=1}^{k} \hat{\lambda}_{(j)}^{(l)} \geq 1- \epsilon,
\end{equation}
where $\epsilon$ is a threshold hyperparameter, usually can be set to $0.01, 0.05, or 0.1$.
\end{enumerate}

\subsection{Non Iterative QCD (NIQCD)}

The estimated number of components, location, and scale parameters in NIQCD are obtained using the steps (1) and (2) in IQCD. That is, $\hat{m} = \hat{m}^{(0)}$, $\hat{\mu}_{l} = \hat{\mu}_{l}^{(0)}$, $\hat{\sigma}_{k} = \hat{\sigma}_{k}^{(0)}$ for $k = 1,\ldots, \hat{m}$.

\begin{enumerate}[leftmargin=1.5em]
\item \textbf{Estimate weight parameters}. 
We estimate the weight parameter $\bm{\lambda}$ by solving $\hat{m}$ linear equations with constraints
\begin{equation}
    \begin{aligned}
        \M{A}\bm{\lambda} = \bm{b}; \textit{s.t.} \sum_{l=1}^{\hat{m}}\lambda_{l}= 1, \bm{\lambda}\geq 0,
    \end{aligned}
\end{equation}

where $b_{l} = F_{n}(\hat{\mu}_{l})$, $\M{A} = [(a_{lk})]$ is an $\hat{m}\times \hat{m}$ matrix and $a_{lk} = G\left((\hat{\mu}_{l}-\hat{\mu}_{k})/\hat{\sigma}_{k}\right)$.

\item \textbf{Update parameters by coordinate descent.\footnote{This step is optional.}} 
The estimated scale parameters could be large on two sides and be small in the middle (high bias).  Define the negative log-likelihood of mixture model: 
\begin{equation}\label{coordinate}
l(\V{\lambda}, \V{\mu}, \V{\sigma}, \tau) = - \log L(\bm{\theta}) = -\sum_{i=1}^{n} \log f_{\hat{m}}(x_i; \V{\lambda}, \V{\mu}, \V{\sigma}).
\end{equation}
Then the coordinate descent method is applied to minimize Equation (\ref{coordinate}) to update the parameters iteratively until they converge. 

\end{enumerate}

\section{Related works}
Here we present two related works and apply it to the Cauchy mixtue model.

\textbf{1. Rousseau and Mengersen (RM)method:} \cite{rousseau2011asymptotic} proved that the posterior behavior of an overfitted mixture model depends on the chosen prior on the proportions $\lambda_j$. They showed that an overfitted mixture model converges to the true mixture, if the Dirichlet-parameters $\lambda_j$ of the prior are smaller than $d/2$.\footnote{$d$ is the dimension of the class-specific parameters, $d = 2$ for mixture of Cauchy} Basically, a deliberately overfitted mixture model with $\hat{m}^{(\text{init})}$, where $\hat{m}^{(\text{init})}$ is usually larger than $m$. A sparse prior with Dirichlet distribution $\lambda_j < 1, \text{ for } j = 1, \ldots, \hat{m}^{(\text{init})}$, on the proportions is then assumed to empty the superfluous classes $(\hat{m}^{(\text{init})} - m)$ during MCMC sampling.

In the overfitted mixture model, each data point is a tuple $(x_i,z_i)$ with $x_i \in \mathbb{R}, z_{i} \in \{1,2,\ldots, \hat{m}^{(\text{init})}\}$ and $z_{i}$ follows a multinomial distribution with $p(z_{i} = k) = \lambda_{k}$ for the simplex $\V{\lambda}$ and represents for the class data where $x_i$ belongs. The joint distribution can be decomposed as $p(x_{i}, z_{i}) = p(x_{i}|z_{i})p(z_{i})$, where $p(x_{i}|z_{i} = k) = g((x_{i}-\mu_k) /\sigma_{k})/\sigma_k$. RM postulates that the observed data is comprised of $\hat{m}^{(\text{init})}$ components with proportions specified by 
$[\lambda_{1},\ldots, \lambda_{\hat{m}^{(\text{init})}}]$. 
We see that $p(x_{i})$ is a mixture model by explicitly writing out this probability: 
$$p(x_{i}) = \sum_{k=1}^{\hat{m}^{(\text{init})}} p(x_{i}|z=k)p(z_{i}=k) = \sum_{k=1}^{\hat{m}^{(\text{init})}}\lambda_{k} g\left({\frac{x_{i}-\mu_k}{\sigma_k}}\right){\frac{1}{\sigma_k}}.$$
Then the corresponding likelihood of the mixture model is
$p(\V{x}) = \prod_{i = 1}^{n} p(x_{i}) = \prod_{i=1}^{n} \sum_{k=1}^{\hat{m}^{(\text{init})}} p(x_{i}|z=k)p(z_{i}=k).$
\cite{nasserinejad2017comparison} claimed a class empty if the number of observations assigned to that latent class is smaller than a certain proportion of the observations in the data set , e.g. $\psi$. In other words, the true number of non-empty classes $m$ could be computed in each MCMC iteration as $\hat{m}^{(l)} = \hat{m}^{(\text{init})} - \sum_{k = 1}^{\hat{m}^{(\text{init})}} I \{n_k^{(l)} \leq n\psi\}$,
where $\hat{m}^{(l)}$ is the estimated number of components in the $l$-th iteration of MCMC sampling, $n_k^{(l)}$ is the number of observations allocated to class $k$ at iteration $l$, $n$ is the sample size and $I(\cdot)$ is the indicator function. $\psi$ is the threshold which can be set to a predefined value, e.g. 0.01, 0.02, or 0.05. Then one can derive the number of non-empty classes $\hat{m}$ based on the posterior mode of the number of non-empty classes based on MCMC iterations.

\textbf{2. Posterior Distribution of Deviance (PDD) method:} An alternative way is to use the posterior distribution of the deviance \cite{aitkin2015new}, by substituting $T$ random draws $\{{\theta_1}, {\theta_2},...,{\theta_T}\}$ from the posterior distribution of model parameter ${\bm{\theta}}$ into the deviance $D(\bm{\theta}) = -2 \log L(\bm{\theta})$, where $L(\bm{\theta}) = \prod_{i=1}^{n}f_{\hat{m}}(x_i;\bm{\theta})$ is the likelihood. Models are compared for the stochastic ordering of their posterior deviance distributions. 

A random variable $X$ is stochastically less than another random variable $Y$ if $F_x(a)\geq F_y(a)$, $\forall a \in (-\infty, \infty)$, with a strict inequality for at least one $a$, where $F_x(\cdot)$ and $F_y(\cdot)$ are CDFs of $X$ and $Y$ respectively. If the CDF of the posterior deviance distribution for model $\mathcal{M}_1$ is stochastically less than that of model $\mathcal{M}_2$, we can say that model $\mathcal{M}_1$ fits the data better than model $\mathcal{M}_2$. 

The CDFs of posterior deviances distribution for different models are compared initially by graphing them in the same plot. It may happen that the CDF curves cross, which indicates that the deviances for competing models are not stochastically ordered. 

In this case, we can take the \textit{most often best} criteria proposed by \cite{aitkin2015new} to compare competing models. Let $s= 1,\ldots, \hat{m}^{(\text{init})}$ represents the index of models. At the $t$-th deviance draw from each model we have deviances $D_1^t,...,D_{\hat{m}^{(\text{init})}}^t$, $t=1,...,T$, where $T$ is the number of samples drawn from each posterior sample. 

For each $t$, we define $\hat{m}^t= \underset{1\leq k\leq \hat{m}^{(\text{init})}}{\arg\min}{\{D^t_k\}}$. Then $\{\hat{m}\}_{t=1}^{T}$ can be viewed as $T$ samples from posterior distribution of $\hat{m}$ who is a discrete random variable taking values on $\{1,2,...,\hat{m}^{(\text{init})}\}$. The \textit{best model} is the one with the largest frequency in the $T$ samples. We call this criteria \textit{most often best} criteria.

\begin{figure}
  \centering
  \includegraphics[scale=.4]{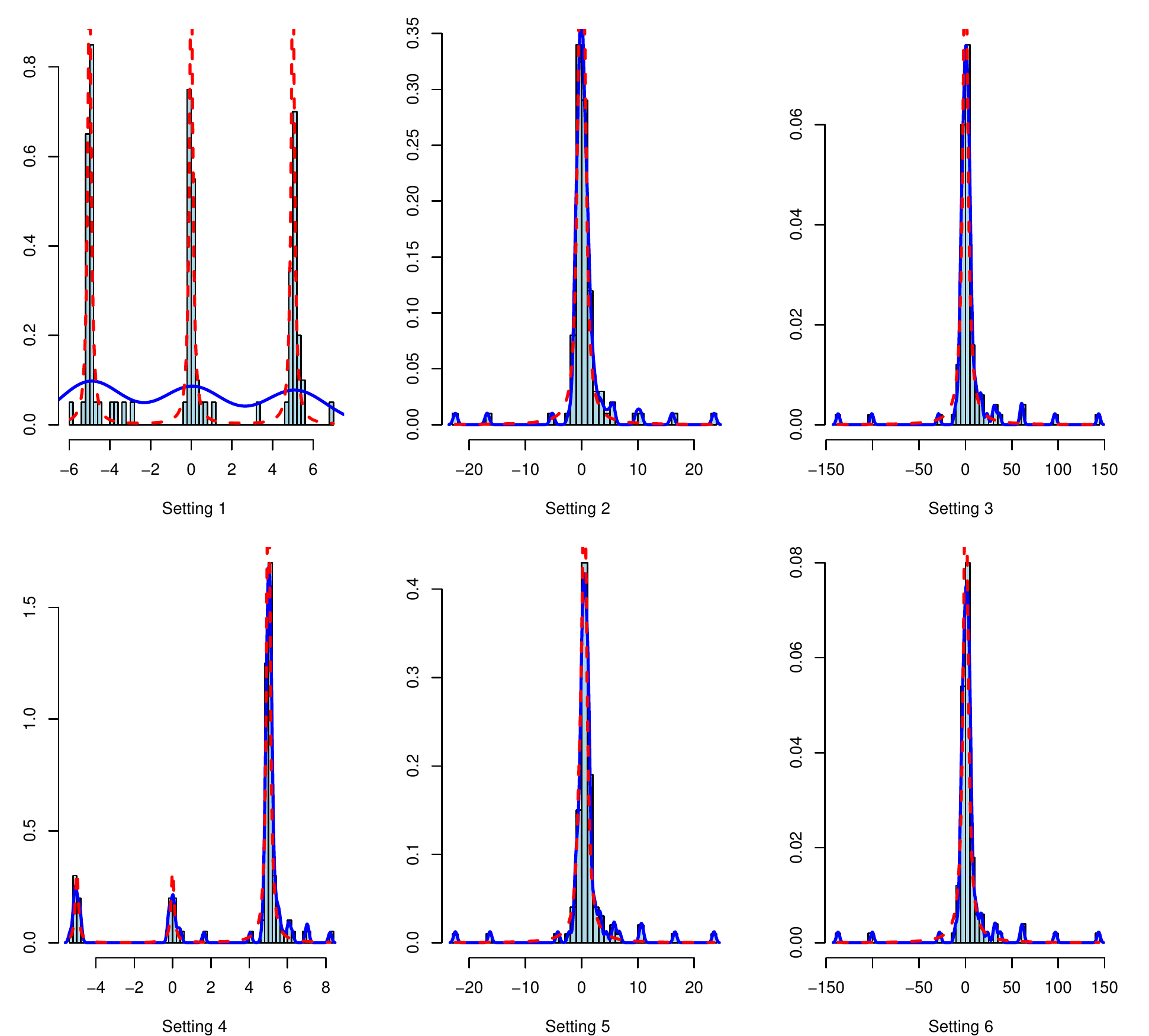}
  \Description{Histograms of randomly selected synthetic data sets with sample size $n = 100$, where the blue solid lines and the red dashed lines represent empirical densities and true densities respectively. The corresponding wDOL is from left to right, from top to bottom (0.01, 0.5, 0.89), (0, 0.09, 0.12) and the corresponding rBCD is (0.1083, 0.0024, 0.0004), (0.1083, 0.0024, 0.0004).}
  \caption{Histograms of randomly selected synthetic data sets with sample size $n = 100$, where the blue solid lines and the red dashed lines represent empirical densities and true densities respectively. The corresponding wDOL is from left to right, from top to bottom (0.01, 0.5, 0.89), (0, 0.09, 0.12) and the corresponding rBCD is (0.1083, 0.0024, 0.0004), (0.1083, 0.0024, 0.0004).}
  \label{Fig: OVL}
\end{figure}

\section{Simulation Study}
\begin{figure*}
  \centering
  \includegraphics[scale=.47]{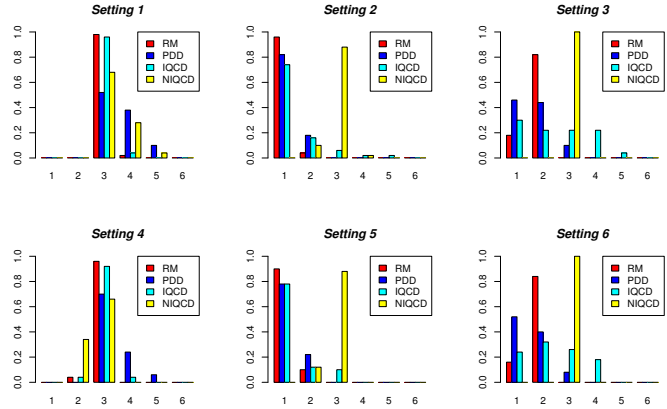}
  \Description{Sample size $n = 100$, given 50 data sets sampled from the mixture density, the correct detection rate. Setting 1 - Setting 6 are from left to right and from top to bottom corresponding. The yellow bar (NIQCD) is consistently good at S2, S3, S5, S5.  NIQCD achieving the true detection rate is over $90\%$. At S1 and S4, NIQCD achieves around $70\%$ and $65\%$ respectively. RM, PDD, IQCD perform not well at S2, S3, S5, S6 and nearly 0 correct detection rate.}
  \caption{Sample size $n = 100$, given 50 data sets sampled from the mixture density, the correct detection rate. Setting 1 - Setting 6 are from left to right and from top to bottom corresponding. The yellow bar (NIQCD) is consistently good at S2, S3, S5, S5.  NIQCD achieving the true detection rate is over $90\%$. At S1 and S4, NIQCD achieves around $70\%$ and $65\%$ respectively. RM, PDD, IQCD perform not well at S2, S3, S5, S6 and nearly 0 correct detection rate.}
  \label{fig:HIST-100}
\end{figure*}
\begin{figure*}
  \centering
  \includegraphics[scale=.47]{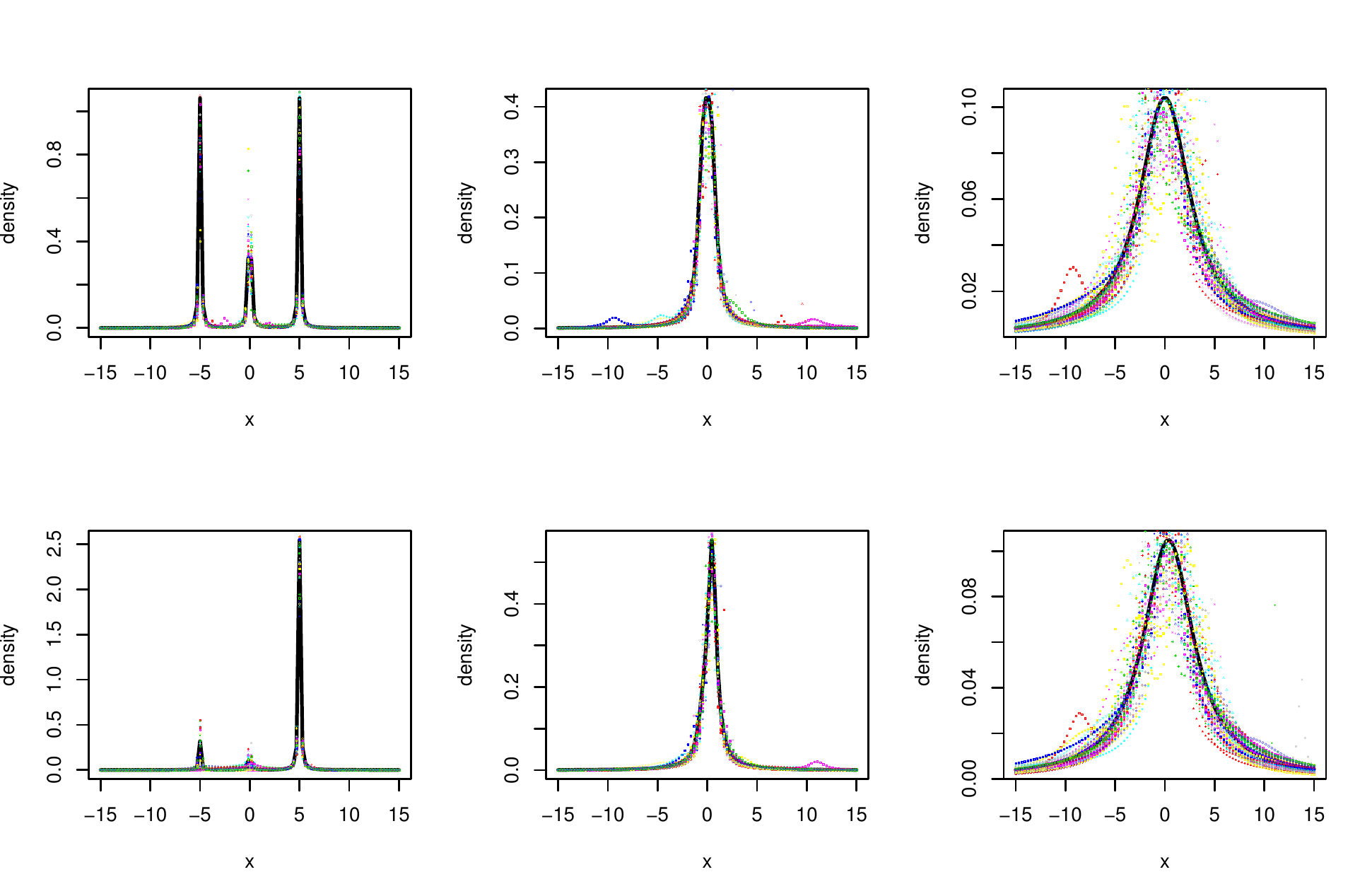}
  \Description{Sample size $n = 100$, given 50 data sets sampled from the mixture density, the fitted lines used the estimated components, location, scale parameters in Cauchy mixture model. Setting 1 - Setting 6 are from left to right and from top to bottom corresponding. At S1 and S4, NIQCD both give the almost perfect fit though under the condition of unequal weighted scenario. At S2, S3, S5, and S6, NIQCD performs not bad at density fitting, which are close to the true density.}
  \caption{Sample size $n = 100$, given 50 data sets sampled from the mixture density, the fitted lines used the estimated components, location, scale parameters in Cauchy mixture model. Setting 1 - Setting 6 are from left to right and from top to bottom corresponding. At S1 and S4, NIQCD both give the almost perfect fit though under the condition of unequal weighted scenario. At S2, S3, S5, and S6, NIQCD performs not bad at density fitting, which are close to the true density.}
  \label{fig:Fit-100}
\end{figure*}
\begin{table*}[t!] 
  \centering
  \begin{minipage}{.5\linewidth} 
  \centering
    \begin{tabular}{c|cccc}
    \hline \hline
      \multirow{2}{*}{Setting}  & \multicolumn{4}{c}{Time[s] (se)}    \\ \cline{2-5}
                                  & RM     & PDD      & IQCD  & NIQCD  \\ \hline
        \multirow{2}{*}{S1}       & 177.01 & 3591     & 105.84 & 0.32    \\ 
                                  & (6.96) & (1161)   & (4.63) & (0.22)  \\ \cline{2-5}
        \multirow{2}{*}{S2}       & 177.55 & 1840     & 93.82  & 0.36    \\ 
                                  & (5.75) & (446)    & (5.65) & (0.13)  \\ \cline{2-5}
        \multirow{2}{*}{S3}       & 167.32 & 775      & 95.71  & 0.28    \\ 
                                  & (2.36) & (227)    & (4.59) & (0.09)  \\ \cline{2-5}
        \multirow{2}{*}{S4}       & 176.94 & 3020     & 102.57 & 0.47    \\ 
                                  & (9.14) & (1451)   & (4.07) & (0.23)  \\ \cline{2-5}
        \multirow{2}{*}{S5}       & 172.72 & 2858     & 89.18  & 0.33    \\ 
                                  & (5.31) & (769)    & (4.79) & (0.18)  \\ \cline{2-5}
        \multirow{2}{*}{S6}       & 166.96 & 830      & 96.11  & 0.28    \\ 
                                  & (2.27) & (283)    & (3.37) & (0.09)  \\ \hline \hline
  \end{tabular}
  \end{minipage}%
  \begin{minipage}{.5\linewidth}
  \centering
    \begin{tabular}{c|cccc}
    \hline \hline
      \multirow{2}{*}{Setting}  & \multicolumn{4}{c}{P-value (se)}  \\ \cline{2-5}
                                  & RM     & PDD       & IQCD   & NIQCD  \\ \hline
        \multirow{2}{*}{S1}       & 0.000  & 0.624     & 0.448   & 0.934    \\ 
                                  & (0.000)& (0.437)   & (0.299) & (0.125)  \\ \cline{2-5}
        \multirow{2}{*}{S2}       & 0.177  & 0.618     & 0.013   & 0.910    \\ 
                                  & (0.206)& (0.346)   & (0.016) & (0.184)  \\ \cline{2-5}
        \multirow{2}{*}{S3}       & 0.007  & 0.516     & 0.037   & 0.918    \\ 
                                  & (0.007)& (0.379)   & (0.035) & (0.165)  \\ \cline{2-5}
        \multirow{2}{*}{S4}       & 0.012  & 0.621     & 0.138   & 0.612    \\ 
                                  & (0.052)& (0.394)   & (0.092) & (0.323)  \\ \cline{2-5}
        \multirow{2}{*}{S5}       & 0.110  & 0.609     & 0.024   & 0.893    \\ 
                                  & (0.170)& (0.373)   & (0.038) & (0.210)  \\ \cline{2-5}
        \multirow{2}{*}{S6}       & 0.006  & 0.547     & 0.040   & 0.880    \\ 
                                  & (0.007)& (0.371)   & (0.050) & (0.205)  \\ \hline \hline
   \end{tabular}  
  \end{minipage}
  \caption{Left: CPU Time (seconds) and corresponding \textit{se} for each scenario with $n=100$. Right: P-value of AD Test and corresponding \textit{se} of P-value for each scenario with $n=100$.}
  \label{tab:table1}
\end{table*}
\begin{table*}
\centering
\begin{tabular}{c|c|cccccc}
\cline{1-8} \hline \hline
Sample size                & Criteria        & S1       & S2     & S3     & S4        & S5     & S6      \\ \hline
\multirow{3}{*}{$n = 100$} & Dectection Rate & RM       & NIQCD & NIQCD & RM        & NIQCD & NIQCD  \\   
                           & GOF by ADT      & NIQCD   & NIQCD & NIQCD & NIQCD    & NIQCD & NIQCD  \\ 
                           & CPU Time        & NIQCD   & NIQCD & NIQCD & NIQCD    & NIQCD & NIQCD  \\ \hline
\multirow{3}{*}{$n = 1000$}& Dectection Rate & RM, IQCD& NIQCD & NIQCD & RM, IQCD & NIQCD & NIQCD  \\
                           & GOF by ADT      & NIQCD   & NIQCD & NIQCD & NIQCD    & NIQCD & NIQCD  \\
                           & CPU Time        & NIQCD   & NIQCD & NIQCD & NIQCD    & NIQCD & NIQCD  \\ \hline \hline
\end{tabular}
\caption{Upper: Sample size $n=100$, given 50 data sets from a mixture density, each cell represents the best method by that row criteria. Two methods simultaneously showing in one cell indicates that the two methods are comparable. Lower: The same law just with different sample size $n=1000$. At S2, S3, S5, S6, the NIQCD is dominant under both sample sizes. At S1, S4, RM is the best detection method but with low detection speed.}
\label{tab: table3}
\end{table*}
In this section, we use $g(z) = 1/[\pi(1+z^2)]$ for Cauchy mixture or any other density s.t. $g(0) \geq g(z) \geq 0, \forall z \in \mathbb{R}$ and $\int_{-\infty}^{\infty} g(z)dz = 1$. Besides, suppose there are $m$ components in the mixture model and $x_{i} \stackrel{iid}{\sim} f_{m}(x), i = 1,2, \ldots, n$.

\subsection{Design}
We compare the performance of these four methods, RM, PDD, IQCD, and NIQCD on simulated data. To evaluate the performance more objectively, we generate synthetic data with two sample sizes, $n = 100$ and $n = 1000$, from $f_{m}(x)$ with $m = 3$ components. We consider 6 different settings summarized below in our experiments with 50 synthetic data sets from each setting. For $n=100$, the histograms of randomly selected synthetic data sets from each setting is present in figure (\ref{Fig: OVL}).

\noindent \textbf{Equal weighted settings:}
\vspace{-1mm}
\begin{itemize}[leftmargin=1.5em]
  \item \textit{Setting 1}: High separation (wDOL = $0.01$, rBCD = $0.1083$): $m = 3$ classes with $(\mu_1,\mu_2, \mu_3) = (-5, 0, 5)$, $(\sigma_1,\sigma_2, \sigma_3) = (0.1, 0.1, 0.1)$, $(\lambda_1,\lambda_2, \lambda_3) = (0.33, 0.33, 0.34)$.

  \item \textit{Setting 2}: Medium separation (wDOL = $0.50$, rBCD = $0.0024$): $m = 3$ classes with $(\mu_1,\mu_2, \mu_3) = (-0.5, 0, 0.5)$, $(\sigma_1,\sigma_2, \sigma_3) = (0.5, 0.5, 0.5)$, $(\lambda_1,\lambda_2, \lambda_3) = (0.33, 0.33, 0.34)$.

  \item \textit{Setting 3}: Low separation (wDOL = $0.89$, rBCD = $0.0004$): $m = 3$ classes with $(\mu_1,\mu_2, \mu_3) = (-0.5, 0, 0.5)$, $(\sigma_1,\sigma_2, \sigma_3) = (3, 3, 3)$, $(\lambda_1,\lambda_2, \lambda_3) = (0.33, 0.33, 0.34)$.
\end{itemize}
\noindent \textbf{Unequal weighted settings:}
\vspace{-1mm}
\begin{itemize}[leftmargin=1.5em]
  \item \textit{Setting 4}: High separation (wDOL = $0.01$, rBCD = $0.1083$): $m = 3$ classes with $(\mu_1,\mu_2, \mu_3) = (-5, 0, 5)$, $(\sigma_1,\sigma_2, \sigma_3) = (0.1, 0.1, 0.1)$, $(\lambda_1,\lambda_2, \lambda_3) = (0.2, 0.3, 0.5)$.

  \item \textit{Setting 5}: Medium separation (wDOL = $0.50$, rBCD = $0.0024$): $m = 3$ classes with $(\mu_1,\mu_2, \mu_3) = (-0.5, 0, 0.5)$, $(\sigma_1,\sigma_2, \sigma_3) = (0.5, 0.5, 0.5)$, $(\lambda_1,\lambda_2, \lambda_3) = (0.2, 0.3, 0.5)$.

  \item \textit{Setting 6}: Low separation (wDOL = $0.89$, rBCD = $0.0004$): $m = 3$ classes with $(\mu_1,\mu_2, \mu_3) = (-0.5, 0, 0.5)$, $(\sigma_1,\sigma_2, \sigma_3) = (3, 3, 3)$, $(\lambda_1,\lambda_2, \lambda_3) = (0.2, 0.3, 0.5)$.
\end{itemize}

\subsection{Prior Distributions and Initial Values}
\begin{itemize}[leftmargin=1.5em]
  \item RM: Vague priors for the location and scale parameters $a_{r} = 0, b_{r}^2 = 10^{3}, c_{r} = 0, d_{r} = 10, \alpha_{r}^{k} = 0.1$, for $k = 1,\dots, \hat{m}^{(\text{init})}_{r}$, where $\hat{m}^{(\text{init})}_{r} = 10$.

  \item PDD: Truncated uniform prior on $\V{\mu}$ and $\V{\sigma}$ and Dirichlet prior on $\V{\lambda}$ are used in PDD method. The location and scale parameters $a_{p} = -20, b_{p} =20, c_{p} = 0, d_{p} =10, \alpha_{p}^{k} = 0.8$, for $k = 1,\dots, \hat{m}^{(\text{init})}_{p}$, where $\hat{m}^{(\text{init})}_{p} = 6$.

  \item IQCD: Initialize the number of components $\hat{m}^{(\text{init})}_{IQCD} = 10$, the cutoff value $\epsilon = 0.05$ (default).
  
  \item NIQCD: Initialize the number of components $\hat{m}^{(\text{init})}_{NIQCD} = 10$, the relative error value is $\kappa = 0.001$ (default).
\end{itemize}
\subsection{Results}
For a small sample size $n = 100$, four methods, RM, PDD, IQCD, and NIQCD, are applied to estimate parameters in the mixture of Cauchy. In Figure (\ref{fig:HIST-100}), the histogram indicates that the ratio of each method correctly chooses over 50 times for each dataset. At S2, S3, S5, S6,  NIQCD is consistently more accurate than other three methods when $m = 3$. 

At Setting 1 (upper left) in Figure (\ref{fig:HIST-100}), two-sample binomial tests are conducted, given critical value of 0.05, corresponding to $Z_{\alpha} = -1.645$, between RM$\sim$NIQCD, PDD$\sim$NIQCD, and IQCD$\sim$NIQCD. The P-Value results are 0.0001, 0.1052, 0.0005 respectively, which means that NIQCD is less effective than RM and IQCD under Setting 1. The possible reason is that NIQCD is very sensitive to abnormal values and allocates more weights on these small components. At Setting 4 (bottom left) in Figure (\ref{fig:HIST-100}), the corresponding P-values are 0.0003, 0.3639, 0.0024, which also shows that NIQCD is not so effective compared to other methods. But it does not bad at detecting the true number of components, which achieves around 0.7 at Setting 1 and Setting 4. 

In Table (\ref{tab:table1}) left, we show the computation time. NIQCD is around 500 times faster than RM, $10^{4}$ times faster than PDD and 300 times faster than IQCD. 

Table (\ref{tab:table1}) right shows the goodness of fit by Anderson Darling test (ADT) and the P-value of NIQCD is around 0.9 at all settings. RM method performs not good at Setting 2 and Setting 5 and P-value is around 0.1 for other settings. PDD's P-value is around 0.6 at all settings. IQCD's P-value ranges from 0.01 to 0.45. 

In Figure (\ref{fig:Fit-100}), 50 estimated density lines of NIQCD method and the corresponding true density has been compared. The black line is the true density, settings from left to right and from top to bottom are Setting 1 to Setting 6 . We see that estimated density lines (colored dashed lines) capture the density line very well although there are few deviance.

In Table (\ref{tab: table3}) top, $n = 100$, we combine three different criteria which are usually used together to evaluate one method, \textit{Detection Rate (DR)}, \textit{Goodness of Fit by ADT}, and \textit{CPU Time}. Under these criteria, NIQCD shows 16 times in overall 18 cells, around 89\% better than other methods under different settings and criteria.

As sample size increases from $n = 100$ to $n = 1000$, in Figure (\ref{fig:HIST-1000}) in Appendix \ref{App: Sec_B}\footnote{For time saving, we only obtained the posterior samples from 1 to 6 Cauchy mixture components to compare for PDD method and simulated it just on $n = 100$}, NIQCD is the best among all settings which achieves above 90\% accuracy over 50 data sets. In Table (\ref{tab:table2}) upper in Appendix \ref{App: Sec_B}, NIQCD is around 500 times faster than the second place IQCD. In Table (\ref{tab:table2}) bottom, NIQCD consistently shows great fit of the data. In Figure (\ref{fig:Fit-1000}) in Appendix \ref{App: Sec_B}, it shows 50 estimated density lines of NIQCD (colored dashed) and the true density is the black line. Compared with Figure (\ref{fig:HIST-100}), with sample size increasing, the goodness of fit becomes much better and with less noise.
\section{Real Data Analysis}
\subsection{Data Description}
Returns from financial assets are normally distributed underpin many traditional financial theories \cite{aparicio2001empirical}, but the reality is that many asset returns do not conform to this law. Empirical distributions fitted on historical data exhibit that higher peaks and heavier tails returns are mostly clustered within a small range around the mean and extreme moves occur more frequently than a normal distribution would suggest. Robert Shiller \cite{grossman1980determinants} observed that fat-tailed distribution can help explain larger return fluctuations. 

In this section, we empirically evaluate our algorithm on the real data from S\&P 500 index between July 1st 2016 and July 1st 2018 with the number of transaction days $n = 503$. We use $r_{t}$ to denote the log return on day $t$ that is defined as $r_t = \ln(x_{t}/x_{t-1})$, where $x_{t}$ and $x_{t-1}$ are closed stock price for day $t$ and $t-1$, respectively. 

Skewness is a measure of distortion or asymmetry of a distribution with a number of 0 indicates complete symmetry.\footnote{Sample Skewness: $b_{1} =\frac{1}{n}\sum_{i=1}{n}(x_{i} - \bar{x})^3/ [\frac{1}{n-1}\sum_{i = 1}^{n} (x_{i} - \bar{x})]^{\frac{3}{2}}$} The sample skewness of the data is $-1.324$, implying that the underlying distribution is slightly skewed to the left and a mixture model could be a reasonable choice to estimate it. On the other hand, kurtosis refers to the degree to which a distribution is more or less peaked than a normal distribution whose kurtosis is 3.\footnote{Sample Kurtosis: $g_2 = \frac{\frac{1}{n}\sum_{i=1}{n}(x_{i} - \bar{x})^4}{ [\frac{1}{n}\sum_{i = 1}^{n} (x_{i} - \bar{x})]^{2}} - 3$} A sample kurtosis of $9.008$ from the data justifies the use of heavy-tailed distribution as the components of the mixture model. Therefore, the mixture of Cauchy model was fitted to the data using NIQCD.
\subsection{Estimated Density}
\begin{figure}
  \centering
  \includegraphics[scale=.45]{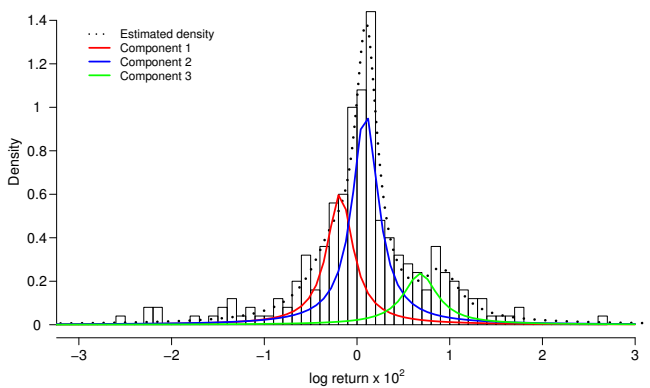}
  \Description{Histogram of log return rate amplified 100 times. The black dashed curve is the estimated density based on NIQCD. Three components are denoted by the red, blue and the green curves.}
  \caption{Histogram of log return rate amplified 100 times. The black dashed curve is the estimated density based on NIQCD. Three components are denoted by the red, blue and the green curves.}
  \label{fig:stock_hist_predict}
\end{figure}

\begin{figure}[t]
  \centering
  \includegraphics[scale=.4]{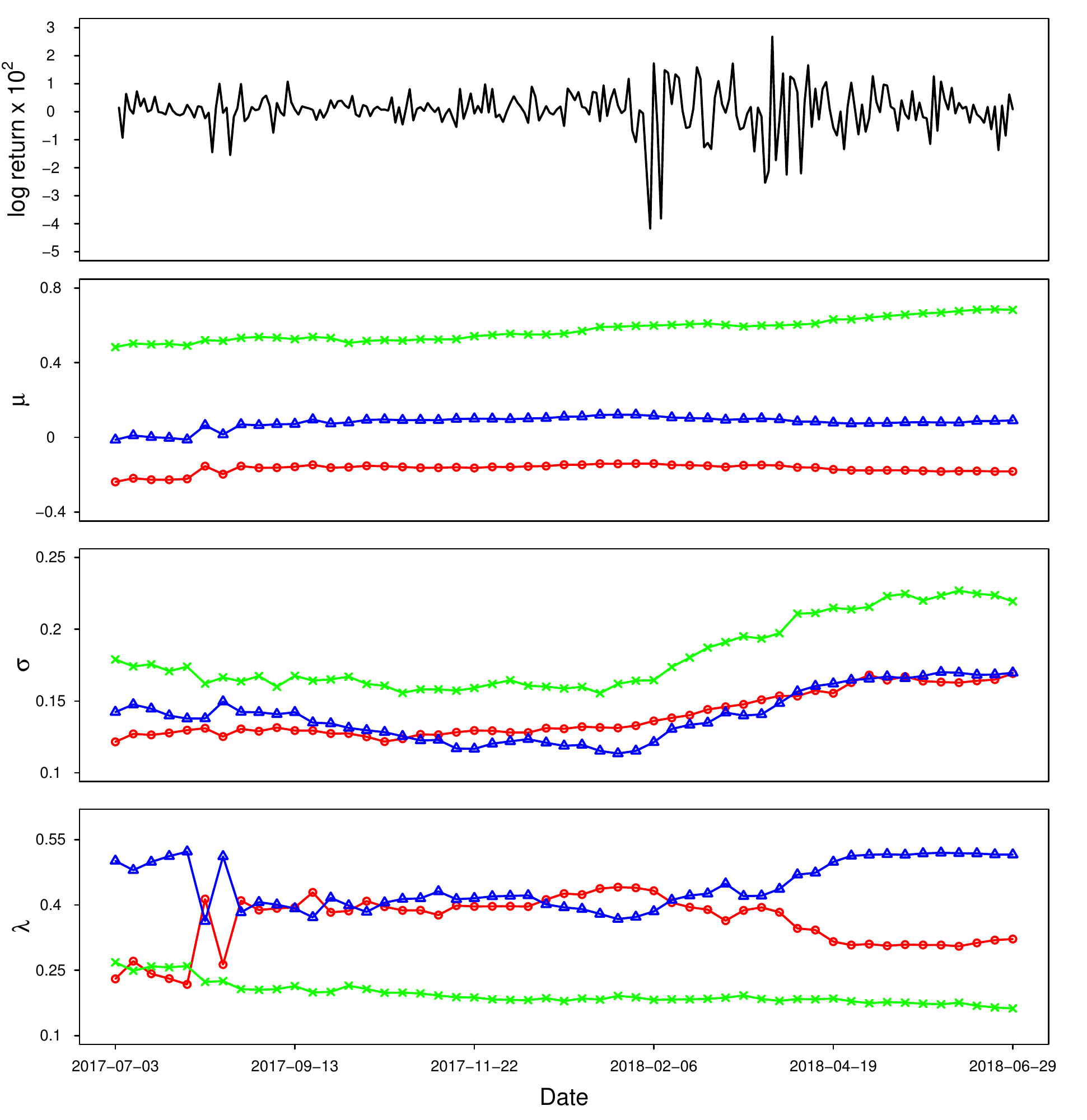}
  \Description{Weekly updated parameters from 07/01/2017 to 07/01/2018. The series of updated parameters of the bear, neutral and bull markets are denoted in red, blue and green respectively. The top panel shows the log return amplified 100 times during this period.}
  \caption{Weekly updated parameters from 07/01/2017 to 07/01/2018. The series of updated parameters of the bear, neutral and bull markets are denoted in red, blue and green respectively. The top panel shows the log return amplified 100 times during this period.}
  \label{fig:stock_week}
\end{figure}

From the results obtained by NIQCD, the optimal number of components for the two years' data is $\hat{m} = 3$. The corresponding estimated weights, locations and scales for each component are shown in \ref{tab:est}. The Anderson Darling Test was performed and the P-value of $0.36$ verifies that the fitted model matches the observations very well.
\begin{table}[H]
\begin{tabular}{l|cccc}
\hline\hline
Parameters & $\hat{\mu}$ & $\hat{\sigma}$ & $\hat{\lambda}$  \\ \hline
Component 1 (bear market) & -0.18 & 0.17 & 0.32 \\ \hline 
Component 2 (neutral market) & 0.09 & 0.17 & 0.52  \\ \hline
Component 3 (bull market) & 0.68 & 0.22 & 0.16  \\ \hline \hline
\end{tabular}
\caption{Estimation of location, scale, and weight parameters}
\label{tab:est}
\end{table}
\vspace{-5mm}
The estimated number of components by NIQCD is in good agreement with the economical common knowledge that  the stock market can be classified into three categories ``bear market'', ``neutral market'' and ``bull market''. In figure (\ref{fig:stock_hist_predict}), we plot the fitted densities of each component. The component 1 (\textit{red} curve) represents the bear market since its average log return rate is negative, the component 2 (\textit{blue} curve) represents the neutral market with the log return rate not far away from zero, and the component 3 (\textit{green} curve) corresponds to the bull market whose log return rate is positive.
The scale parameters of the bull market (0.22) is slightly larger than the other two markets, which means that the bull market is more volatile. The estimated weight of the neutral market (0.52) shows that it occupies this market for a longer time during these two years compared to the bear market and the bull market, which meets the common knowledge that the market is usually at the neutral state. The possible reason for the relative large weight (0.32) of the bear market is that the market declined a lot and exhibited dramatic volatility from Feb 2018 to Apr 2018 as shown in Figure (\ref{fig:stock_week}) (top).
To test the robustness of our algorithm, we update the estimated parameters weekly from July 2017 in an offline fashion as shown in Figure (\ref{fig:stock_week}). That is, the training data accumulate weekly and are fitted with NIQCD as a whole. The second panel of Figure (\ref{fig:stock_week}) indicates that the locations of the three market didn't change significantly but the bull market did show slightly positive shift. From the updated scale parameters in the third panel, the volatility was relatively stable before Feb 2018 but increased considerably after that for all markets, which was probably caused by the huge fluctuation in technology stocks. The $\lambda$ part Figure (\ref{fig:stock_week}) shows the bull market moved down a little while the bear market moved up, implying that the whole market changed from relatively weak bear market to neutral market.
\subsection{Prediction of Return Category}
We use the NIQCD method to get corresponding estimated parameters and then apply the three components' densities to calculate the corresponding probabilities from July 1, 2018 to August 1, 2018 with 20 transaction days. Then we get the index of each day with respect to the largest probability and assign category '-1' to component 1, category '0' to component 2, and category '1' to component 3, 
\begin{equation*}
    \begin{aligned}
    \text{Category}_{t} = 
        \begin{cases}
        -1, &\text{component 1's density is largest at day} \hspace{0.1cm} t  \\
        0, &\text{component 2's density is largest at day} \hspace{0.1cm} t \\
        1, &\text{component 3's density is largest at day} \hspace{0.1cm} t
        \end{cases}
    \end{aligned}
\end{equation*}
\begin{table}[t]
\centering
\begin{tabular}{cc|ccc}
\hline \hline
    Date       & Category & Date        & Category \\ \hline 
    2018-07-03 & -1       & 2018-07-18  & -1\\            
    2018-07-05 & 1        & 2018-07-19  & 0 \\
    2018-07-06 & 0        & 2018-07-20  & -1\\
    2018-07-09 & 1        & 2018-07-23  & 0 \\
    2018-07-10 & 0        & 2018-07-24  & -1\\
    2018-07-11 & 1        & 2018-07-25  & 0 \\
    2018-07-12 & 0        & 2018-07-26  & -1\\
    2018-07-13 & 1        & 2018-07-27  & 0 \\
    2018-07-16 & -1       & 2018-07-30  & 0 \\ 
    2018-07-17 & 0        & 2018-07-31  & -1\\ \hline \hline
\end{tabular}
\caption{The prediction of category of return from July 1 2018 to August 1 2018. ``-1'' represents that point lies in red line. ``0'' represents that point lies in blue line and ``1'' represents that point lies in green line.}
\label{tab4:stock_pred}
\end{table}
The result is shown in Table (\ref{tab4:stock_pred}). There are 7 days belonging to component 1, 9 days belonging to component 2, 4 days belonging to component 3, which indicates a shift from a neutral market to a bear market during this period.

\section{Conclusion}
Our study is a proof-of-concept study and provides a straightforward option for parameter estimation in mixture models with heavy-tailed components. Note that our algorithm is not limited to Cauchy mixture models, and can also be applied to other densities. The current work is not designed to be exhaustive in incorporating all potential mixture methods. We present the simulation results with six settings and prove that our algorithm show great performance even in the most difficult separation settings (Setting 2, Setting 3, Setting 5, Setting 6). Besides, our algorithm has relatively small computational complexity compared to other methods, the extension of NIQCD to an online version should not be difficult.

\bibliographystyle{ACM-Reference-Format}
\bibliography{ref}

\appendix
\clearpage

This appendix provides detailed supplementary information for the NIQCD algorithm and additional simulation results. Readers may refer to the publicly-available code for more implementation details.\footnote{https://github.com/Likelyt/Mixture-Model-tools.}

\section{Algorithm}
The detailed NIQCD algorithm is shown below.
\begin{algorithm}
  \caption*{\textbf{Algorithm} \algnameNIQCD} \label{alg:algnameNIQCD}
  \begin{algorithmic}[1]
    \STATE {\bfseries Initialize:} {Given data $x$, iteration $L$, threshold $\epsilon$, a rough estimate of components $\hat{m}^{\text{(init)}}$}
    \STATE {Initial estimate location parameters: $$\hat{\mu}_{k}^{(0)} = x_{([\frac{nk}{\hat{m}^{\text{(init)}}+1}])}, k = 1, \ldots, \hat{m}^{\text{(init)}}$$}
    \STATE {Number of components determination: Then use change point method to detect the sudden change of $\hat{\V{\mu}}^{(0)}$ and get the new estimated $\hat{m}^{(0)}$.}
    \STATE {Initial estimate scale parameters: $$\hat{\sigma}_{k}^{(0)} = \frac{x_{([\frac{n(k+1)}{\hat{m}^{(0)}+2}])} - x_{([\frac{nk}{\hat{m}^{(0)}+2}])}}{2\tau}, k = 1, \ldots, \hat{m}^{(0)}$$}
    \STATE {Initial estimate weight parameters: $$\V{\lambda} = (\M{A}\Tra \M{A})^{-1} \M{A}\Tra \V{b}$$
    \STATE {Rescale weight parameters: $$\hat{\V{\lambda}}^{(0)} = \frac{\V{\lambda}}{\sum_{k = 1}^{\hat{m}^{(0)}} \lambda_{k}}$$}}
    \STATE {After getting the initial estimated parameters $(\hat{\V{\lambda}}^{(0)}, \hat{\V{\mu}}^{(0)}, \hat{\V{\sigma}}^{(0)})$. Use coordinate descent algorithm to find the minimum of the log-likelihood.}
    \FOR{$l=1$ {\bfseries to} $L$ }
    \STATE {$\hat{\V{\sigma}}^{(1)} = \underset{\V{\sigma}}{\arg\min}\; L(\V{\sigma}| \hat{\V{\lambda}}^{(0)}, \hat{\V{\mu}}^{(0)}, x)$}
    \STATE {$\hat{\V{\mu}}^{(1)} = \underset{\V{\mu}}{\arg\min}\; L( \V{\mu} |\hat{\V{\lambda}}^{(0)}, \hat{\V{\sigma}}^{(1)}, x)$}
    \STATE {$\hat{\V{\lambda}}^{(1)} = \underset{\V{\lambda}}{\arg\min}\; L(\V{\lambda}| \hat{\V{\mu}}^{(1)}, \hat{\V{\sigma}}^{(1)}, x)$ \\
      $s.t. \hspace{2mm} \V{\lambda} \geq 0$, $\sum_{k = 1}^{m} \lambda_{k} = 1$}
    
    \IF{$|L(\hat{\V{\lambda}}^{(l)}, \hat{\V{\mu}}^{(l)}, \hat{\V{\sigma}}^{(l)}, x) - L(\hat{\V{\lambda}}^{(l-1)}, \hat{\V{\mu}}^{(l-1)}, \hat{\V{\sigma}}^{(l-1)}, x)|$ $\leq \kappa |L(\hat{\V{\lambda}}^{(l-1)}, \hat{\V{\mu}}^{(l-1)}, \hat{\V{\sigma}}^{(l-1)}, x)|$}
    \STATE {Break}
    \ENDIF
    \ENDFOR
    \STATE {Get final estimated parameters: $(\hat{\V{\lambda}}^{(l)}, \hat{\V{\mu}}^{(l)}, \hat{\V{\sigma}}^{(l)}, \hat{m})$}
  \end{algorithmic}
\end{algorithm}

\newpage 
\section{Additional Simulation Results}
\label{App: Sec_B}
Additional simulation results for sample size $n=1000$ are shown below including CPU Time, P-value of AD Test, estimation of number of components, and fitting lines for 50 data sets. 
\begin{table}[h] 
  \centering
    \begin{tabular}{c|cccc}
    \hline \hline
      \multirow{2}{*}{Setting}  & \multicolumn{4}{c}{Time[s] (se)}     \\ \cline{2-5}
                                  & RM      & PDD   & IQCD   & NIQCD   \\ \hline
        \multirow{2}{*}{S1}       & 2021    &  -    & 522     & 0.96     \\ 
                                  & (51.81) &  -    & (9.00)  & (0.41)   \\ \cline{2-5}
        \multirow{2}{*}{S2}       & 2263    &  -    & 505     & 1.24     \\ 
                                  & (118.95)&  -    & (12.47) & (0.17)   \\ \cline{2-5}
        \multirow{2}{*}{S3}       & 2073    &  -    & 500     & 1.50     \\ 
                                  & (60.80) &  -    & (9.81)  & (2.14)   \\ \cline{2-5}
        \multirow{2}{*}{S4}       & 2010    &  -    & 505     & 2.33     \\ 
                                  & (88.32) &  -    & (7.82)  & (1.39)   \\ \cline{2-5}
        \multirow{2}{*}{S5}       & 2200    &  -    & 493     & 1.34     \\ 
                                  & (107.80)&  -    & (8.38)  & (0.29)   \\ \cline{2-5}
        \multirow{2}{*}{S6}       & 2084    &  -    & 496     & 1.26     \\ 
                                  & (48.91) &  -    & (7.34)  & (0.64)   \\ \hline \hline
  \end{tabular}
\end{table}

\begin{table}[h]
  \centering
    \begin{tabular}{c|cccc}
    \hline \hline
      \multirow{2}{*}{Setting}  & \multicolumn{4}{c}{P-value (se)}   \\ \cline{2-5}
                                  & RM      & PDD    & IQCD    & NIQCD \\ \hline
        \multirow{2}{*}{S1}       & 0.952   &  -     & 0.002    & 0.959   \\           
                                  & (0.067) &  -     & (0.002)  & (0.083) \\ \cline{2-5}
        \multirow{2}{*}{S2}       & 0.000   &  -     & 0.000    & 0.928   \\ 
                                  & (0.001) &  -     & (0.000)  & (0.132) \\ \cline{2-5}
        \multirow{2}{*}{S3}       & 0.000   &  -     & 0.000    & 0.949   \\ 
                                  & (0.000) &  -     & (0.000)  & (0.072) \\ \cline{2-5}
        \multirow{2}{*}{S4}       & 0.000   &  -     & 0.000    & 0.108   \\ 
                                  & (0.000) &  -     & (0.000)  & (0.282) \\ \cline{2-5}
        \multirow{2}{*}{S5}       & 0.002   &  -     & 0.000    & 0.941   \\ 
                                  & (0.014) &  -     & (0.000)  & (0.120) \\ \cline{2-5}
        \multirow{2}{*}{S6}       & 0.000   &  -     & 0.000    & 0.952   \\ 
                                  & (0.000) &  -     & (0.000)  & (0.067) \\ \hline \hline
   \end{tabular}  
  \caption{Upper: \textit{CPU Time (seconds) and corresponding \textit{se} for each scenario with $n=1000$.} Lower: \textit{P-value of AD Test and corresponding \textit{se} of P-value for each scenario with $n=1000$.}}
  \label{tab:table2}
\end{table}

\newpage
\begin{figure*}
  \centering
  \includegraphics[scale=.475]{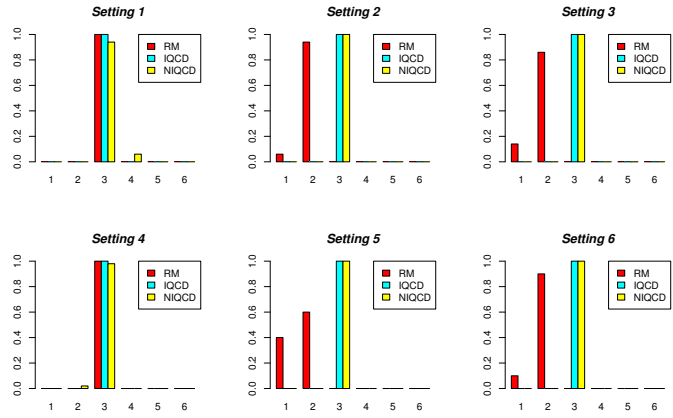}
  \Description{Histogram of the number of components detected, sample size n = 1000.}
  \caption{Histogram of the number of components detected, sample size n = 1000.}
  \label{fig:HIST-1000}
\end{figure*}

\begin{figure*}
  \centering
  \includegraphics[scale=.4755]{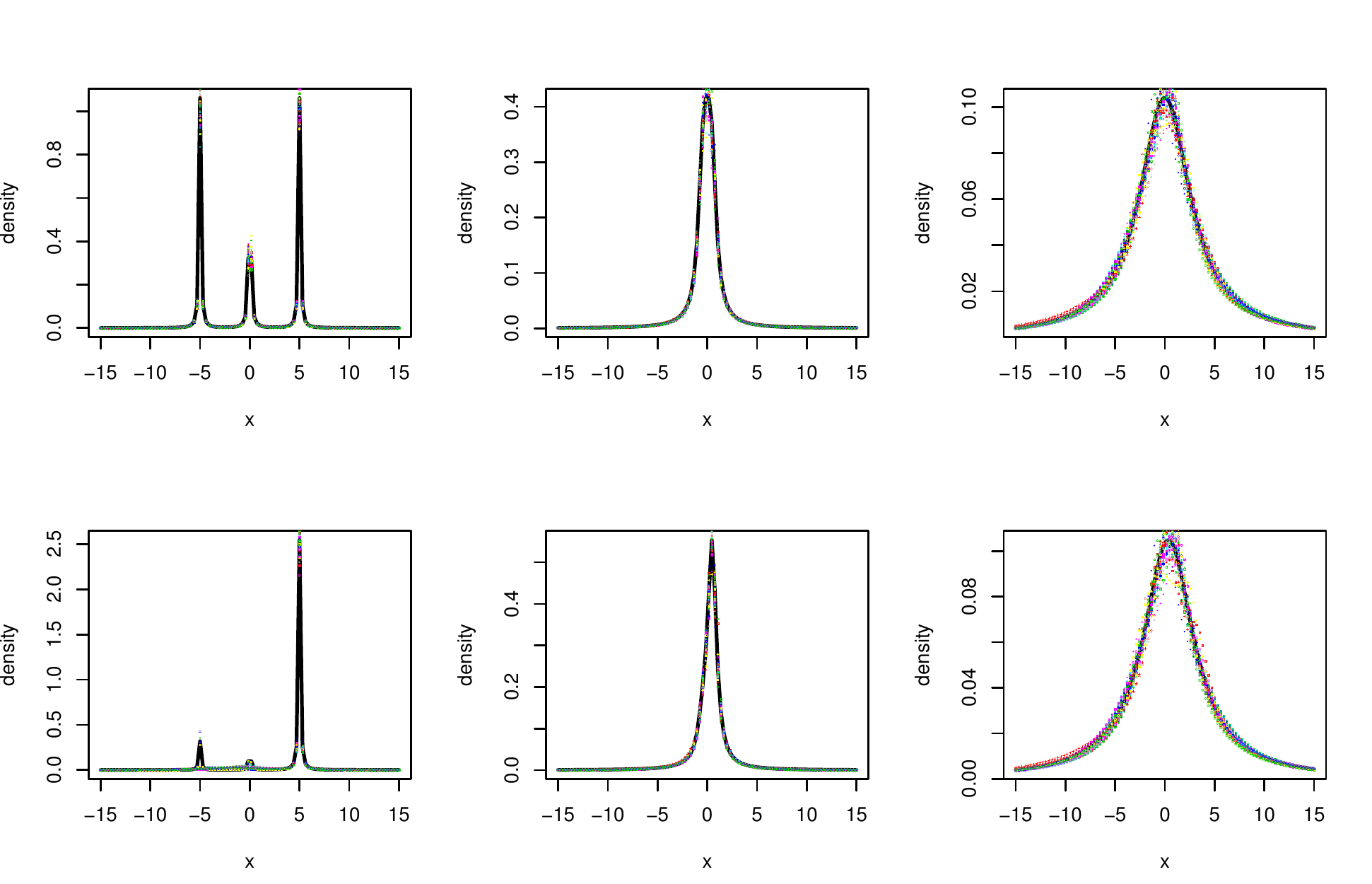}
  \Description{Fitted dashed lines by NIQCD method with 50 datasets sampled from true mixture model, sample size n = 1000.}
  \caption{Fitted dashed lines by NIQCD method with 50 datasets sampled from true mixture model, sample size n = 1000.}
  \label{fig:Fit-1000}
\end{figure*}

\end{document}